\documentclass[10pt,twocolumn,letterpaper]{article}

\usepackage{cvpr}              

\usepackage{graphicx}
\usepackage{amsmath}
\usepackage{amssymb}
\usepackage{booktabs}
\usepackage{algpseudocode}

\usepackage[pagebackref,breaklinks,colorlinks]{hyperref}

\usepackage[capitalize]{cleveref}
\crefname{section}{Sec.}{Secs.}
\Crefname{section}{Section}{Sections}
\Crefname{table}{Table}{Tables}
\crefname{table}{Tab.}{Tabs.}

\def\cvprPaperID{*****} 
\def\confName{CVPR}
\def\confYear{2022}

\usepackage{graphicx}
\usepackage{amsmath}
\usepackage{amssymb}
\usepackage{booktabs}
\usepackage{array}
\usepackage{algpseudocode}
\usepackage{algorithm}
\usepackage{algpseudocode}
\usepackage{tabularx}
\usepackage[pagebackref,breaklinks,colorlinks]{hyperref}

\usepackage[capitalize]{cleveref}
\crefname{section}{Sec.}{Secs.}
\Crefname{section}{Section}{Sections}
\Crefname{table}{Table}{Tables}
\crefname{table}{Tab.}{Tabs.}

\def\cvprPaperID{*****} 
\def\confYear{2022}

\begin{document}
\title{Traditional methods in Edge,Corner and Boundary detection }
\author{Sai Pavan Tadem \\ saipavanthadem@gmail.com}

\maketitle

\begin{abstract}
This is a review paper of traditional approaches for edge, corner, and boundary detection methods. There are many real-world applications of edge, corner, and boundary detection methods. For instance, in medical image analysis, edge detectors are used to extract the features from the given image. In modern innovations like autonomous vehicles, edge detection and segmentation are the most crucial things. If we want to detect motion or track video, corner detectors help. I tried to compare the results of detectors stage-wise wherever it is possible and also discussed the importance of image prepossessing to minimise the noise. Real-world images are used to validate detector performance and limitations.
\end{abstract}
\section{Introduction}
\label{sec:intro}
Computer vision is a field of artificial intelligence which extracts a meaning full information from the digital images and responds to a particular situation.AI allows computers to think and act whereas computer vision allows a machine to observe and understand the images.There are many practical applications,for example self driving vehicles,Medical imaging,Industry automation,Robotics etc.We can partition the revolution of computer vision into a three major steps like low level understanding(for example edge detectors,optical flow etc) ,high level understanding(for example face detection and SIFT etc, and era of deep learning(using CNN,RNN,AlexNet for segmentation or classification etc).
The collected data from the cameras or sensors is huge,we don't need every information of it for a specific task,that creates unnecessary complexity.So before performing the computations we need to reduce the information.for example in case of self driving cars the generated  rgb image takes lot of memory and computationally heavy,if we apply edge detectors on it we will get a binary image.Its a very light information so easy for computations,In finger print reader also we don't need additional information like color etc ,we just need edges.\\
Corners are considered as important features of the image because they are invariant to the image translations like rotation,illumination.There are many application with corners,for example motion tracing.Moving object may rotate but the corners wont change.Harris corner detection is one most often used method.
Boundary detection is a complex problem,need higher level computations to find outer layer of the object.For example if we apply edge detection on zebra then it given the edges on the surface(white and black interface) and outer surface but if we apply boundary detection algorithm it just gives outer edge of it.


\section{Image Prepossessing}
If we want to detect the changes or edges in an images,we need to apply find change in direction,this process is more sensitive to the noise present in the images,the entire output is effected due to this.
\begin{figure}[htp]
    \includegraphics[width=8cm, height=4cm]{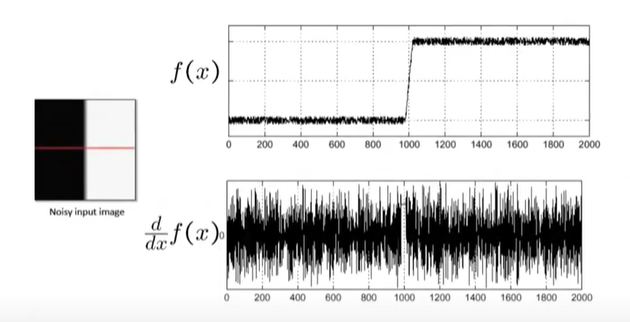}
\end{figure}
So it is important to remove the noise by using some standard noise removal methods.In the below example,the input image consist of noise so after applying differentiation the noise is too high,it is difficult to detect the edge position.

Based on the type of noise present in the image we have to use select the desired filter,Some basic filters are listed below.

\begin{tabularx}{0.4\textwidth} { 
  | >{\raggedright\arraybackslash}X 
  | >{\centering\arraybackslash}X 
  | >{\raggedleft\arraybackslash}X | }
 \hline
  Filter Type  & Advantages & Working Principle  \\
\hline
 Mean Filter & Removes the Gaussian noise & Based on average value of pixel  \\
 \hline
 
\end{tabularx}

\begin{tabularx}{0.4\textwidth} { 
  | >{\raggedright\arraybackslash}X 
  | >{\centering\arraybackslash}X 
  | >{\raggedleft\arraybackslash}X | }
 \hline
 Median Filter  & Removes salt,pepper and speckle noise  & based on median  \\
 \hline
 Wiener Filter  & Removes the blurring from image 22  & Inverse filtering in frequency domain  \\
\hline
 Hybrid Filter & Removes salt,pepper and speckle and blurring & combination of median and wiener filter   \\ 
\hline

\end{tabularx}



\section{Edge detection}
We consider edge is a discontinuity or sudden change in intensity of pixels.It can be surface normal discontinuity,depth discontinuity ,surface color discontinuity and illumination discontinuity.Edges gives the structure of an object,We can imaging a shape of a table,book,horse even without any information of color we can still identify them only with the help of edges.The Idea of edge detection is not only to get the edges from the image,but to create foundations for image matching,shape finding,object detection,segmentation etc.
\subsection{Image Gradients}
The change in pixel intensity is measured with the help of gradients\cite{jacobs2005image}.The image gradient $\nabla$ f=$(\frac{\partial f }{\partial x} , \frac{\partial f }{\partial y})$.It is also important to find the direction and magnitude of change,they are defined as following ways. \\Orientation of edge or gradient direction is defined by using $\theta$ .The value of $\theta$ = $\arctan{\phi}$ ,where the value of $\phi$ is $\frac{\frac{\partial f}{\partial x}}{\frac{\partial f}{\partial y}}$.The magnitude is defined by $\sqrt{{\frac{\partial f}{\partial x}}^2+{\frac{\partial f}{\partial y}}^2}$.

\subsection{Gradient method to detect the edges}
From below figure(2),we can observe the change of intensities  black to white.if we apply derivative on it we can observe peak as shown in figure.If we again apply derivative one more time i.e. second order derivative the edges are converted into an impulse as shown in figure.
Gradient in x direction gives the edges in x and gradient in y direction gives edges in y direction.
\begin{figure}[htp]
    \includegraphics[width=8cm, height=8cm]{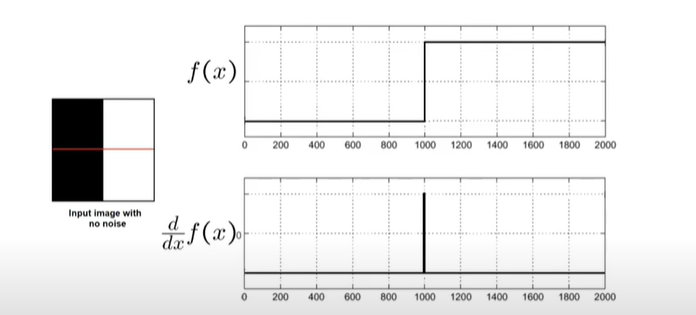}
\end{figure}
\section{Edge detectors}
Edge detection is done with the help of a kernel which moves on the original image and detect the features.There are few different methods.
\subsection{Prewitt Edge Detector}
It is a first order differential operator.The middle or center pixel value is zero and adjecent values are -1 and 1 .The are multiplied with pixel values and difference is used to detect the edge intensity.



\begin{align*}
\begin{bmatrix}
1 & 1 & 1 \\ 
0 & 0 & 0 \\ 
-1 & -1 & -1 \\
\end{bmatrix}
\begin{bmatrix}
-1 & 0 & 1\\
-1 & 0 & 1\\
-1 & 0 & 1\\
\end{bmatrix}
\end{align*}

\subsection{Sobel Edge Detector}
The sobel edge detector\cite{7006020} is used to find the horizontal and vertical edges in the given image,so thats why we are using two kernels or operators.In case of prewitt detector the gradient difference is low but for instance take horizontal operator,from center it has 2,-2 ,so the difference enhances the intensity than the prewitt operator so,the edges are more stronger in sobel filter as compared to prewitt.

\begin{align*}
\begin{bmatrix}
1 & 2 & 1\\
0 & 0 & 0\\
-1 & -2 & -1 \\
\end{bmatrix} 
\begin{bmatrix}
1 & 0 & -1\\
2 & 0 & -2\\
1 & 0 & -1\\
\end{bmatrix} 
\end{align*}

\subsection{Robert Edge Detector}
There is one more edge detector called Robert\cite{7006020},which used to detect the edges in diagonal directions.
\newline

\begin{align*}
\begin{bmatrix}
1 & 0\\
0 & -1\\
\end{bmatrix} \\
\begin{bmatrix}
0 & -1\\
1 & 0\\
\end{bmatrix} 
\end{align*}

here all above methods are to find edges in single directions,but in real case the images have edges i multiple directions so we need a more robust edge detectors which can find the edges in multiple directions.But for that we need good image gradient methods
\subsection{Laplacian Edge Detector}
Sobel and prewitt edge operators uses two kernels with single derivatives.we have to apply the kernels twice.The laplacian is double derivative,It has single kernel to get edge.We just need to apply only one kernel to get the edges
The most frequently used operators are 
\begin{align*}
\begin{bmatrix}
0 & -1 & 0\\
-1 & 4 & -1\\
0 & -1 & 0 \\
\end{bmatrix} 
 \begin{bmatrix}
-1 & -1 &-1\\
-1 & 8&-1\\
-1 & -1 &-1\\
\end{bmatrix} 
\end{align*}

\subsection{Canny Edge Detector}
It is one of the most commanly used edge detector.The issue with sobel filter is thick edges,in practical cases our end goal is to detect the edges so there is no point of representing the edge thickness,Canny edge detector\cite{heath1998comparison} makes edges to more thinner(makes one pixel width).It is ultimately like reduction of unwanted information very much helpful for post processing.The Canny edge detector follows below steps.

\begin{itemize}
    \item Noise reduction:The edges are detected by using gradients,so these gradients are very sensitive to the noise present in the image,It is very important to apply gaussian blurr.
    \item Gradients Finding:By using sobel operators we have to find the edges in horizontal and vertical direction and from there magnitude and theta are calculated.
    \item Non-maximum suppression: Now we have to find the two neighbours in -ve and +ve directions.If the magnitude of the present pixel is lower than the magnitude of the neighbour pixel,then make current pixel to zeoro.Otherwise no changes applied to the present pixel. 
    \item Double threshold: As of now we got thinner edges from the non-maximum suppression but there exist some unwanted edges(like not strong intensive edges) so we can remove them by simply applying the threshold.There are two level of threshloding.The threshold one is less than the second threshold.The gradients that are smaller than the lower threshold are suppressed and the gradients higher than the high threshold are stronger so no changes applied and marked as strong pixels(they are included in the final edge map).The pixels between these two levels are considered as weak pixels so their presence is decided by the hysteresis method.
    \item Edge tracking by hysteresis:Let say w is a week pixel and s is a strong pixel(we marked it in above step like which pixel is week and strong)
    if w has any connection with strong pixel directly(i.e. with any one of its eight neighbours) then this w is considered as strong pixel and added to the final edge map otherwise w is assigned to zero.
    
\end{itemize}
\subsection{Comments on edge detectors}
The below figure shows the results of all edge detectors that we discussed as of now.
\begin{figure}[htp]
    \includegraphics[width=8cm, height=8cm]{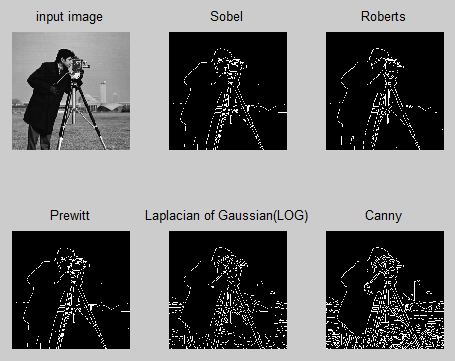}
\end{figure}
Here the results generated from the canny edge detector are very good as compared to the other methods.In case first order derivative approches like sobel,prewitt and Robert edge detectors consisting higher noise than the second order filters.

\section{Corner Detection}

Corner is a point where two edges meets.In case of edge the rapid change happens in one direction but here the rapid change occures in two directions. 
\begin{figure}[htp]
    \includegraphics[width=8cm, height=4cm]{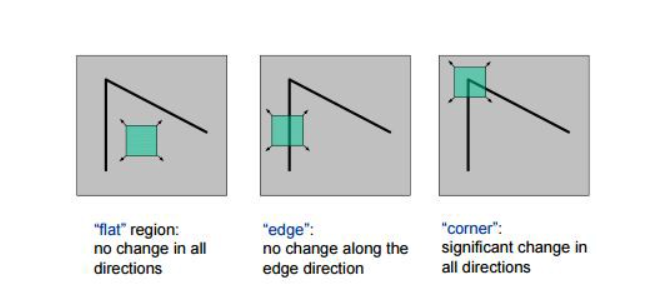}
\end{figure}
We already saw edges and there importance but we wont get all information from edges.For example from the above figure if we place kernel on flat region no changes are observed,even on the edge also no changes but when kernel is at corner some significant changes are observed.we can use this knowledge in  motion detection, image registration, video tracking, image mosaicing, panorama stitching, 3D reconstruction and object recognition.\\ Now let us understand about Eigen values and vectors which plays a very crucial role in detecting the corners.Eigen value tells us the varience of the data along the axis.The derivative of distributions are as below for flat,edge and corner.\\

\begin{figure}[htp]
    \includegraphics[width=9cm, height=4cm]{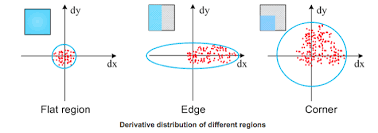}\\
\end{figure}
Here for flat region both the axis are equal in lenghts it means that the varience along dx and dy are same so $\lambda1 and \lambda2$ are also equal and low(because the variences are also low).In case of edge the varience along the dx is high($\lambda1$ is high) and the varience along the dy is low($\lambda2$ is low).Similarly for corner the distribution is spread to the both the axis and almost both eigen values are equal and large. 
\begin{figure}[htp]
    \includegraphics[width=8cm, height=4cm]{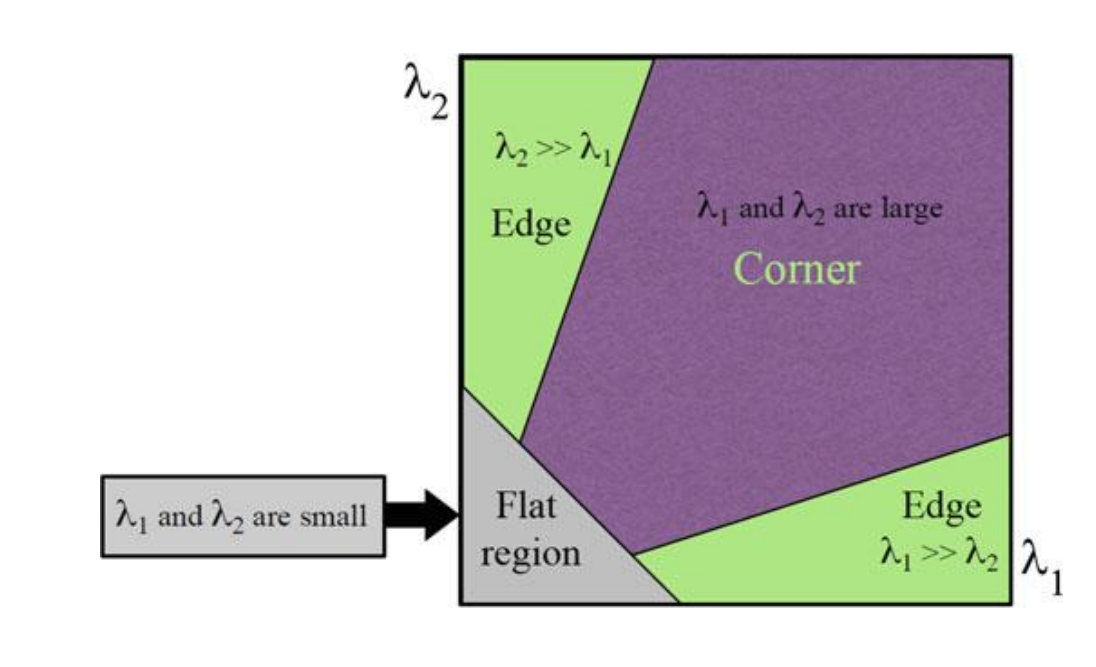}\\
\end{figure}
Now we got the basic idea to detect the corner among all three types.But we need to find the location of edge in the given image,By harris corner detection equation  R=$ \lambda 1 \lambda 2 + K(\lambda 1+ \lambda 2)^2 $,we can achieve it.The response after the corner response is shown below.
\begin{figure}[htp]

    \includegraphics[width=8cm, height=3cm]{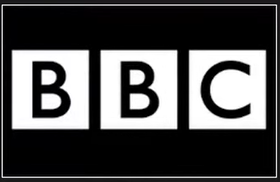}\\
\end{figure}
This equation is combination of eigen values,Initially we set a threshold point for R,if R is higher than the threshold than it is a corner.But still the corner is not perfect because we are taking set of values{all R values higher than threshold}.
\begin{figure}[htp]
    \includegraphics[width=8cm, height=4cm]{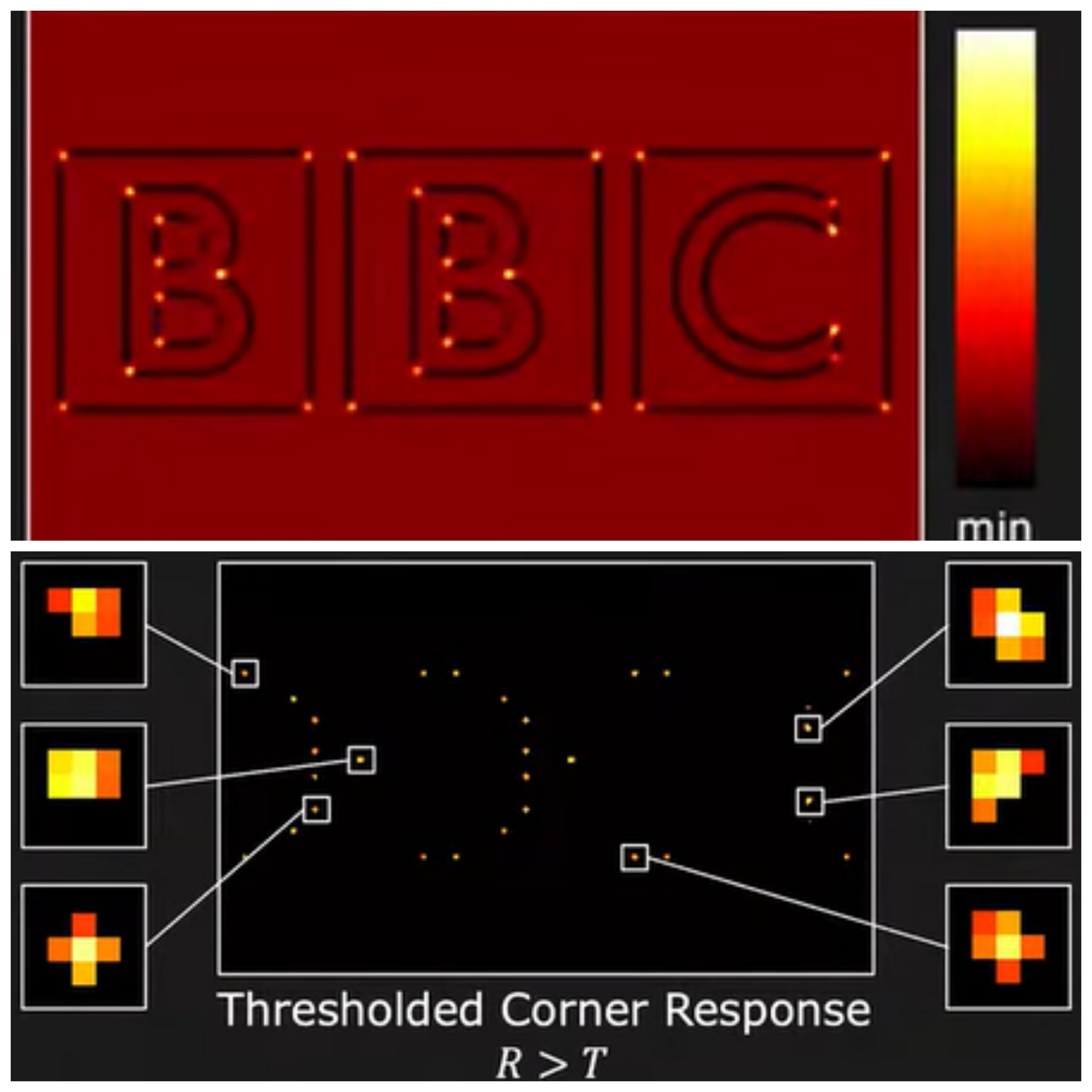}\\
    \caption{Top image-Corner Response R,Bottom Image-Threshold corner response with wide corners}
\end{figure}
The corners looks like below.The corner should be a single pixel values but here we have a group of intensities.
We can remove them by using non maximal suppression(which we already seen in case of canny edge detector),in this method we will pick a pixel and tries to find its neighbour intensities,if current pixel is lower intensity than its neighbour pixel than we make it zero.
\begin{figure}[htp]
    \includegraphics[width=8cm, height=4cm]{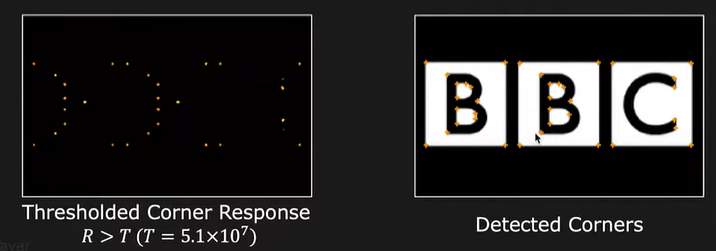}\\
    \caption{Right-Threshold corner response after non maximum suppression,Left-detected corner spots}
\end{figure}

\section{Boundary Detection}
Finding boundary of the object is most important in many applications like segmentation.There are few notable methods to find boundaries,In this paper explained about hough transform.
\subsection{Hough transform}
Hough transform\cite{herout2013review} is a power full approach to detect the boundaries in a binary image.We already seen methods to find edges in a given image,We have to find the boundaries from this edge points.To detect the edges we can use sobel,priwitte,RObert,Laplace,Canny or basically any edge detection method.
\subsubsection{Line Detection}
we have to find edges in a given image by using edge detection methods.Canny edge detection is most widely used method.The result is a binary image which consist of only edges pixels.Now we have to use this edge pixels to detect the boundary of the objects in the image\cite{herout2013review}.
\begin{figure}[htp]
    \includegraphics[width=8cm, height=4cm]{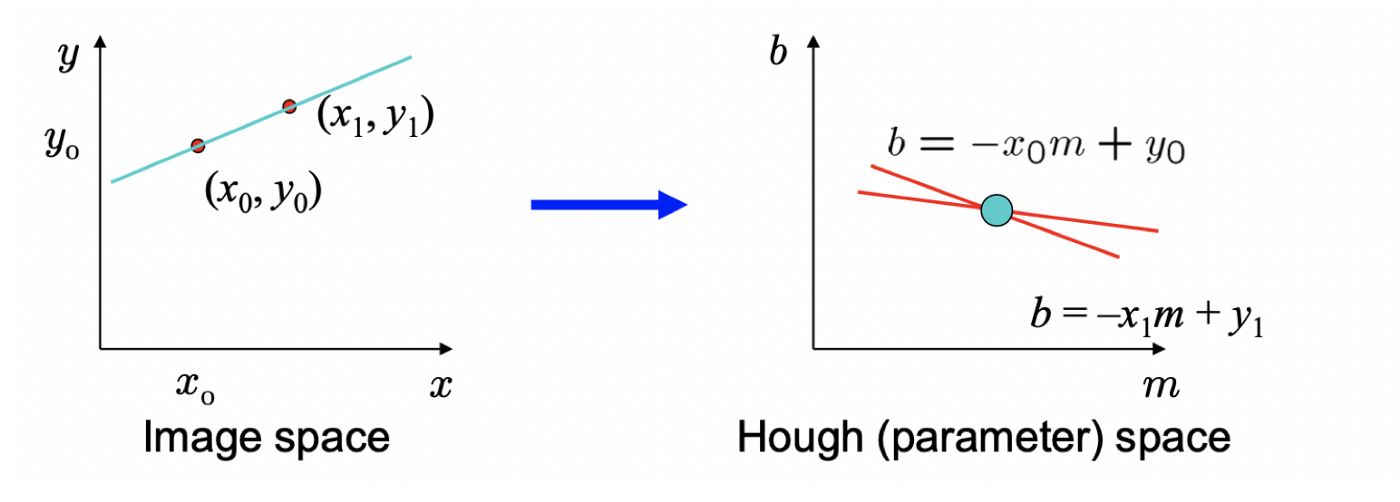}\\
\end{figure}
Consider the below case,let us take one point on the line as $(x\_o,y\_o)$ so we can write the equation as $ y\_o=mx\_o+c $.\\
Every point on the line has same m,c.If we project the same equation on to the parameter space i.e $ c=y_o-mx \_o $.So on C and m plan all line equations derived from the points on image space intersects at same c,m point(i.e because c,m are common for all points on the line on image place).In case of multiple edges in an image as shown in figure there exist multiple group of intersections as shown below.So this way parameter space helps us to detect the lines in a given image.There is a difficulty in this method,the value of slop varies from $-\infty$  to $+\infty$,It means that we need large accumulator,So we need another approach to avoid this.We can convert our line equation to the polar form i.e $ x\cos{\theta}+y \sin{\theta}=\rho $,range of $\theta$ is $0$ to $\pi$ and range of $\rho$ is $0$ to $1$.So w can reduce the accumulator size.
This method creates cosins in the parameter space.\\
Find edges of the original image using edge detectors.We need to create a parameter space $\rho$ and $\theta$ . Initialize an accumulator array A($\rho$,$\theta$) ,all values are zero initially.Now apply loop on edge image and find for the edge(pixel),at the current pixel find all possible $\theta$ and use it to evaluate $\rho$,then after identify index points of $\rho$ and $\theta$ and increment it by one.Now after this apply loop through the updated accumulator and find higher intensity points and those have information about the line. 
\begin{figure}[htp]
    \includegraphics[width=8cm, height=8cm]{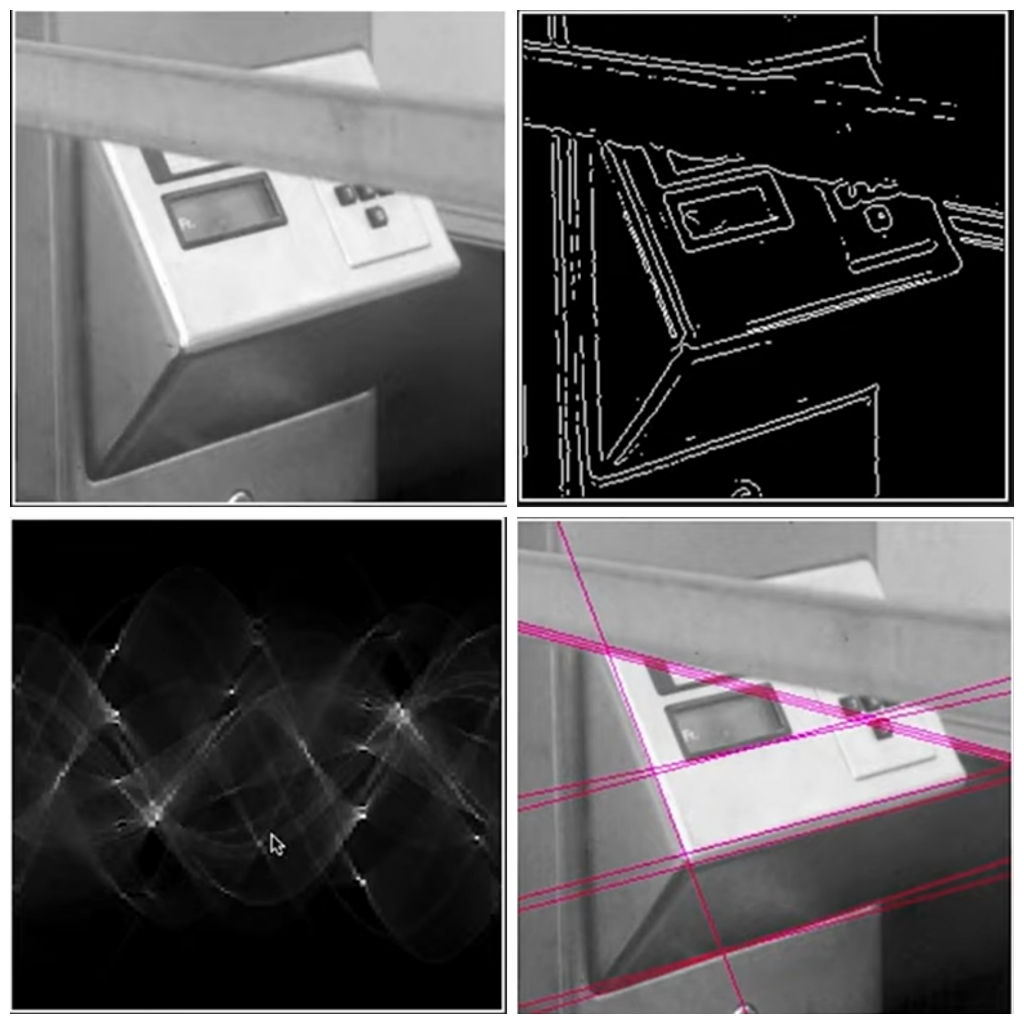}\\
    \title{Line Hough transformed output}
    \caption{Top left - original Image,Top Right - Edge image,bottom left -Output,bottom right - detected lines}
\end{figure}

\subsubsection{Circle Detection}
The idea is same we need to change image space to parameter space,the equation of the circle is $ (x\_o\-a)^2 + (y\_o\-b)^2 = r^2 $,So a,b,r are the parameters.So the equations in parameter space becomes $ a=x\_o +/- r\cos{\theta} ,b=y\_o +/- r\sin{\theta} $.So a circle in image space have a,b,r values common and (x,y) points are changes.In parameter space all these different point of circle circumstance(image space) makes circle with the common a,b,r values.So the corresponding intersection of all circles(common point) is the center of the original circle in image space.\\Lets take coins image and find edges of it.
\begin{figure}[htp]
    \includegraphics[width=8cm, height=5cm]{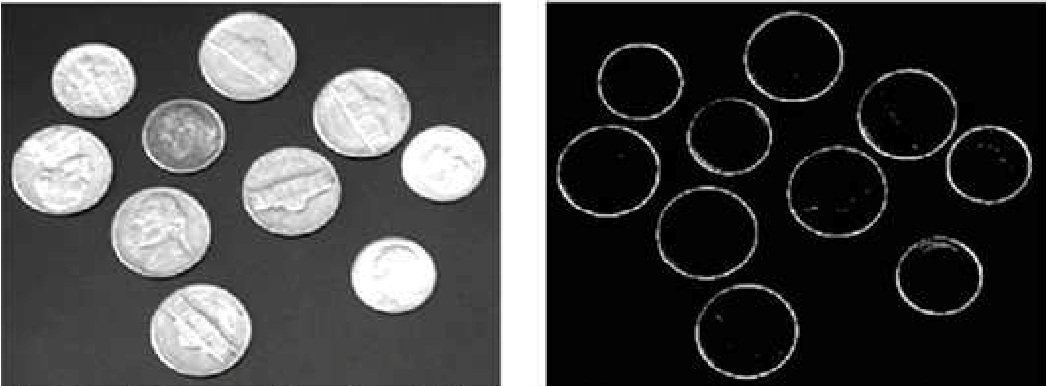}\\
    \title{Lefts side image is Input image and right side image is Sobel edge image}
\end{figure}
If apply the hough transform and with the radius suits for Penny the accumulator detects the penny as shown below.
\begin{figure}[htp]
    \includegraphics[width=8cm, height=8cm]{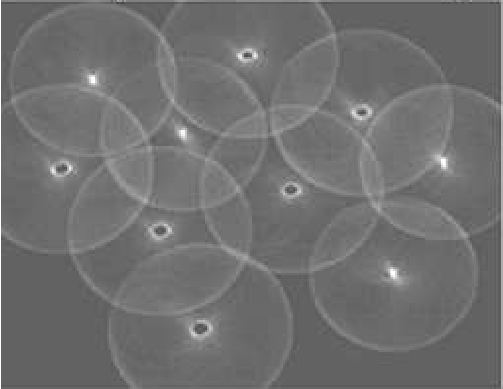}\\
    \title{Circle Hough Transform image(A(x,y)) with radius of 25 pixels}
    
\end{figure}
If we change the value of radius that suits for quarter then it detects them as shown below.
\begin{figure}[htp]
    \includegraphics[width=8cm, height=8cm]{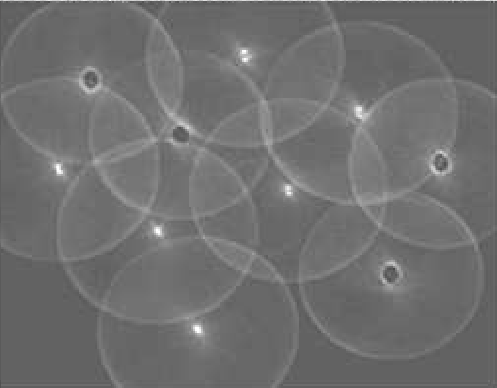}\\
    \title{Circle Hough Transform image(A(x,y)) with radius of 30 pixels}
    
\end{figure}

The hogh transform depends on parameter space,for example it is valid for a line,circle because the have some fixed parameter(they have regular shapes),but what if we have an irregular shape ? It means we cant express those shapes with the generalised expressions.So to over come this there is an another method called Generalised hough transform\cite{BALLARD1981111}.
\subsection{Generalised hough transform}
Consider the below figure and a reference point(x,y) within the object.This reference point should be offline to the points on boundary.All points on edge are called edge points.This shape is irregular so we need a model which represents this shape.The object which cannot be described by the generalised equations are described by the discrete information.It is obtained by the edge pixel location and edge pixel phase as shown in figure(x,y,$\phi$).
\begin{figure}[htp]
    \includegraphics[width=8cm, height=4cm]{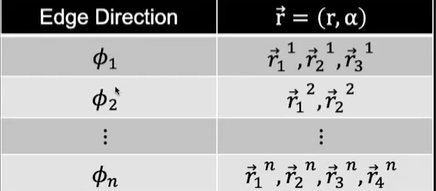}\\
\end{figure}
The above table show the all possible vectors in the directions of $\phi$.Now the model is ready.So our aim is to find the location of the reference point if indeed in the object lied on the image,for that we need to do voting.

 now we have an accumulator matrix with votes we have to find the local maximum value and that point represents the x,y.

 \begin{figure}[htp]
    \includegraphics[width=8cm, height=4cm]{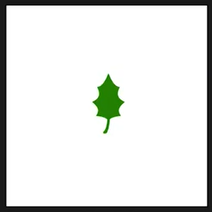}\\
    \caption{model-Image boundary to be detected}
\end{figure}
Consider the below image,which consists of different kind of liefs.Our model is a green lief so accumulator should hold the information about the existence or position of the lief. 
\begin{figure}[htp]
    \includegraphics[width=8cm, height=5cm]{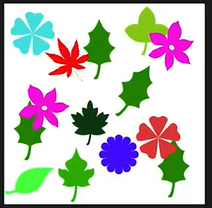}\\
    \caption{image}
\end{figure}
So after applying the algorithm we will get accumulator array A.This array consists maximum intensity at a position of model in image.
\begin{figure}[htp]
    \includegraphics[width=8cm, height=4cm]{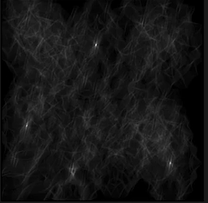}\\
    \caption{ generalised hough transform output i.e A(x,y)}
\end{figure}
\subsubsection{Comments on hough transform}
The Hough transform can be used on disconnected edges because it purely depends on the voting method. It is very sensitive to noise due to the parametric space. It is effective for simple shapes like circles and lines(regular shapes) because they can be expressed in a generalised format. In the case of irregular shapes, there is no generalised expression, so we have to use generalised high transform. The major limitation is that this method is more complex when we have a higher number of parameters.

\section{Observations and Conclusion}
Gradient\cite{Harris1988ACC} provides the edge's location, magnitude, and direction.Laplacian provides only the location of the edge. The Canny edge detector is performing well out of all edge detectors. Harris corner detection is good at finding edges, but it requires you to set a threshold value manually to extract important features, which is mostly done with the trial and error method. So this method won't assure the right threshold value every time. Hough transform is good for regular shapes like lines and circles but does not work for irregular shapes. That's where generalised hough transform comes into the picture. It allows making one model to represent the irregular shape and then following the voting method to detect the desired boundary or shape.

\section{Acknowledgement}
I would like to thank Professor Shree K. Nayar for the free online course on the Internet "First Principles of Computer Vision," which helped me to write this paper.

{\small
\bibliographystyle{ieee_fullname}
\bibliography{egbib}
}

\end{document}